\documentclass[sigconf]{acmart}

\usepackage{color}
\usepackage{subfigure} 
\usepackage{multirow}
\usepackage[linesnumbered,ruled,vlined]{algorithm2e}
\SetKwInput{KwInput}{Input}                
\SetKwInput{KwOutput}{Output}              
\usepackage{epsfig}
\usepackage{balance}

\copyrightyear{2021} 
\acmYear{2021} 
\setcopyright{acmcopyright}
\acmConference[MM '21]{Proceedings of the 29th ACM International Conference on Multimedia}{October 20--24, 2021}{Virtual Event, China}
\acmBooktitle{Proceedings of the 29th ACM International Conference on Multimedia (MM '21), October 20--24, 2021, Virtual Event, China}
\acmPrice{15.00}
\acmDOI{10.1145/3474085.3475553}
\acmISBN{978-1-4503-8651-7/21/10}

\begin{document}
\fancyhead{}

\title{MHFC: Multi-Head Feature Collaboration for Few-Shot Learning}







\author{Shuai Shao$^{\dagger,1}$, Lei Xing$^{\dagger,2}$, Yan Wang$^{3}$, Rui Xu$^{1}$, Chunyan Zhao$^{4}$, Yanjiang Wang$^{*,1}$, Baodi Liu$^{*,1}$}
\thanks{
$^{\dagger}$ indicates equal contribution.
$^{*}$ indicates corresponding author.
}
\affiliation{%
  \institution{$^{1}$College of Control Science and Engineering, China University of Petroleum (East China)}
  \streetaddress{}
  \city{}
  \state{}
  \country{}
}
\affiliation{%
  \institution{$^{2}$College of Oceanography and Space Informatics, China University of Petroleum (East China)}
  \streetaddress{}
  \city{}
  \state{}
  \country{}
}
\affiliation{%
  \institution{$^{3}$Beihang University}
  \streetaddress{}
  \city{}
  \state{}
  \country{}
}
\affiliation{%
  \institution{$^{4}$Suzhou Centennial College}
  \streetaddress{}
  \city{}
  \state{}
  \country{}
}
\affiliation{%
  \institution{shuaishao@s.upc.edu.cn, upc\_xl@163.com, wangyan9509@gmail.com, ruixu@s.upc.edu.cn}
  \streetaddress{}
  \city{}
  \state{}
  \country{}
}
\affiliation{%
  \institution{zhaocy@scc.edu.cn, yjwang@upc.edu.cn, thu.liubaodi@gmail.com}
  \streetaddress{}
  \city{}
  \state{}
  \country{}
}

\renewcommand{\shortauthors}{Trovato and Tobin, et al.}

\begin{abstract}
Few-shot learning (FSL) aims to address the data-scarce problem.
A standard FSL framework is composed of two components: \textbf{(1)} Pre-train. Employ the base data to generate a CNN-based feature extraction model (FEM). \textbf{(2)} Meta-test. Apply the trained FEM to acquire the novel data's features and recognize them. FSL relies heavily on the design of the FEM. However, various FEMs have distinct emphases. For example, several may focus more attention on the contour information, whereas others may lay particular emphasis on the texture information. The single-head feature is only a one-sided representation of the sample.
Besides the negative influence of cross-domain (e.g., the trained FEM can not adapt to the novel class flawlessly), the distribution of novel data may have a certain degree of deviation compared with the ground truth distribution, which is dubbed as distribution-shift-problem (DSP).
To address the DSP, we propose \textbf{Multi-Head Feature Collaboration (MHFC)} algorithm, which attempts to project the multi-head features (e.g., multiple features extracted from a variety of FEMs) to a unified space and fuse them to capture more discriminative information. Typically, first, we introduce a subspace learning method to transform the multi-head features to aligned low-dimensional representations. 
It corrects the DSP via learning the feature with more powerful discrimination and overcomes the problem of inconsistent measurement scales from different head features. 
Then, we design an attention block to update combination weights for each head feature automatically. It comprehensively considers the contribution of various perspectives and further improves the discrimination of features. We evaluate the proposed method on five benchmark datasets (including cross-domain experiments) and achieve significant improvements of \textbf{2.1\%}-\textbf{7.8\%} compared with state-of-the-arts.
\end{abstract}




\maketitle
\begin{figure}[h]
    \centering
    \subfigure[SS-R-Fea]{
    \begin{minipage}[t]{0.3\linewidth}
    	\begin{center}
    		\includegraphics[width=1\linewidth]{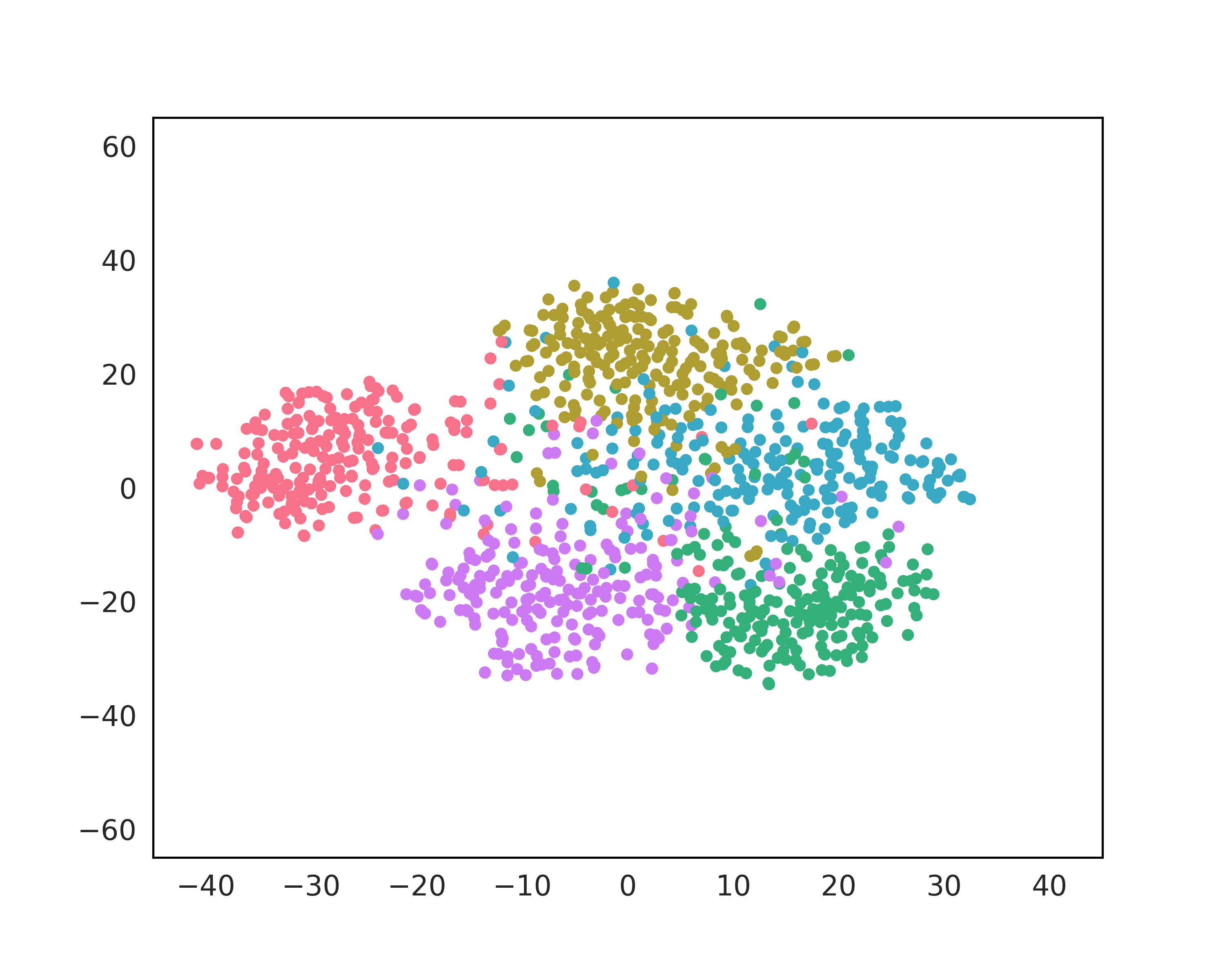}
    	\end{center}
    	\label{fig: SS-R-Fea}
    \end{minipage}%
    }%
    \subfigure[SS-M-Fea]{
    \begin{minipage}[t]{0.3\linewidth}
    	\begin{center}
    		\includegraphics[width=1\linewidth]{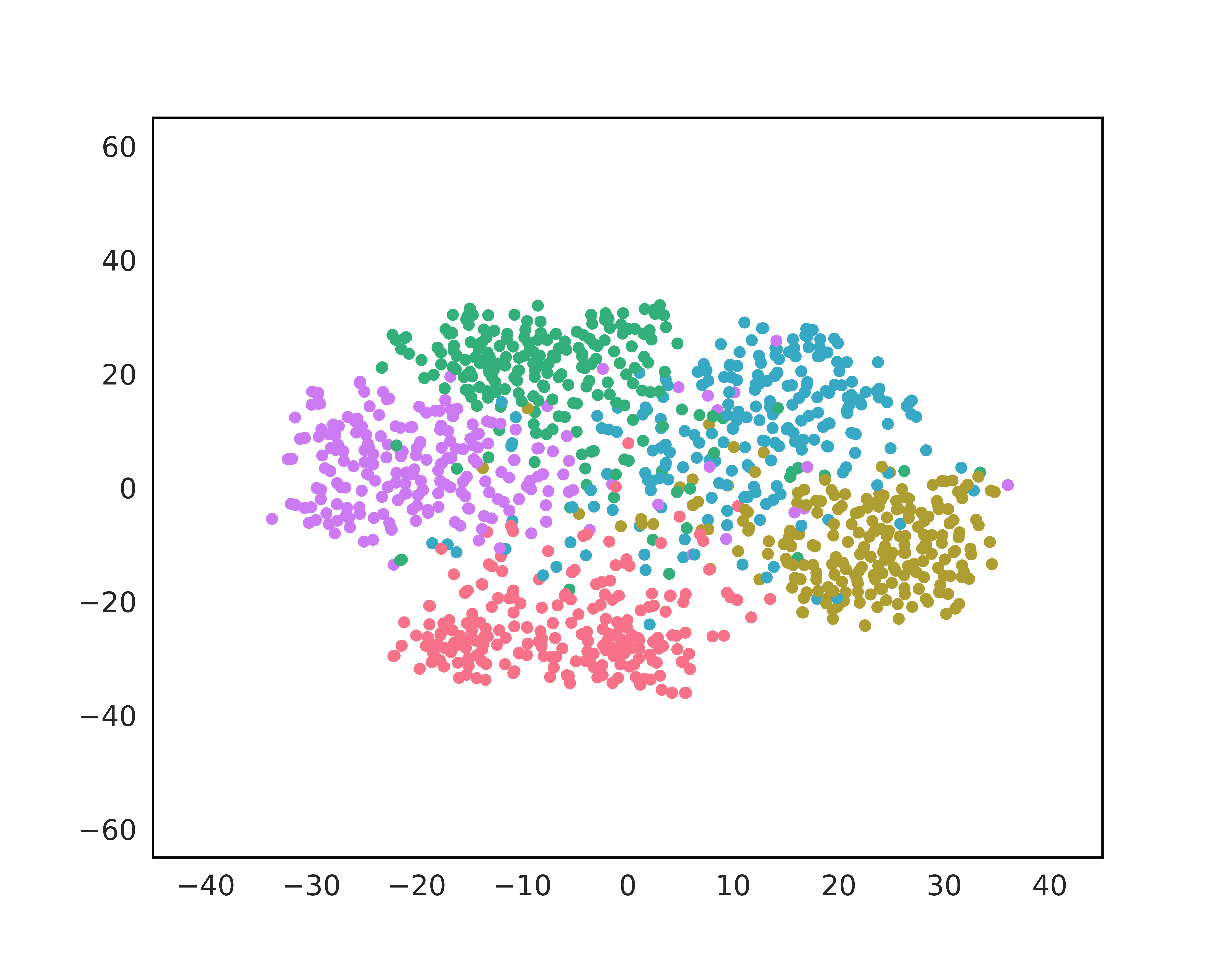}
    	\end{center}
    	\label{fig: SS-M-Fea}
    \end{minipage}%
    }%
        \subfigure[Multi-Head-Fea]{
    \begin{minipage}[t]{0.3\linewidth}
    	\begin{center}
    		\includegraphics[width=1\linewidth]{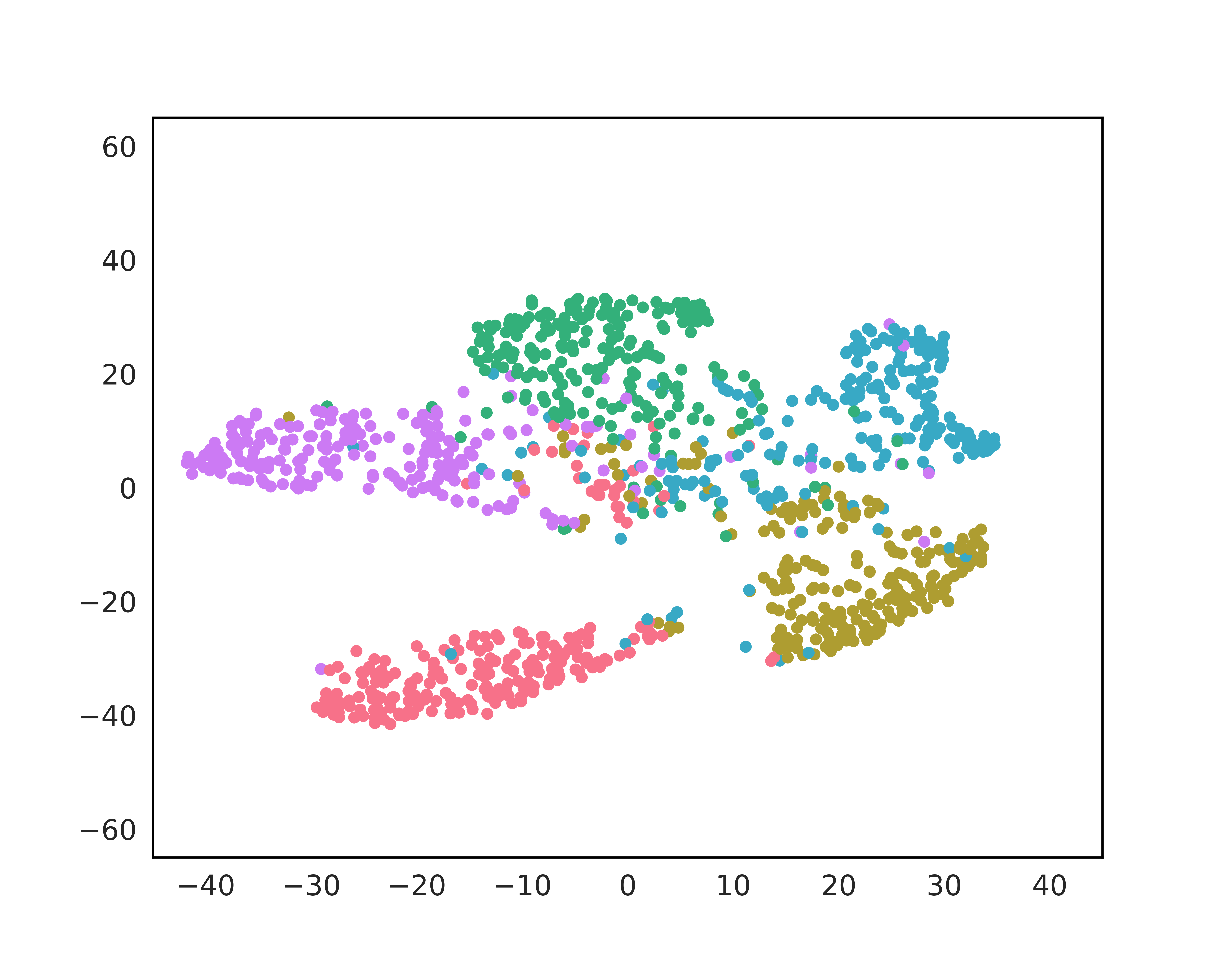}
    	\end{center}
    	\label{fig: Multi-Head-Fea}
    \end{minipage}%
    }%
    \caption{t-SNE visualization of features on mini-ImageNet. "SS-R-Fea" and "SS-M-Fea" indicate two categories of different single-head features (for the details, please refer to \textbf{Section \ref{subsec: Multi-Head Feature Extraction Model}}). "Multi-Head-Fea" denotes the fusion feature after the proposed MHFC. Multi-Head-Fea based distribution is more helpful to classification tasks.}
    \label{fig: tsne}
\end{figure}
\section{Introduction}
\label{sec: introduction}

Machine learning has help machines achieve outstanding performances in computer vision tasks, such as person re-identification \cite{wang2020dense,wang2020unsupervised,fan2020contextual}, image classification \cite{yan2016image,lin2020orthogonalization,shao2020label}. 
One indispensable factor is attributed to the large-scale labeled data.
However, as the limitation of actual circumstances, it may be infeasible to collect large amounts of labeled data in the real world. 
Thus, few-shot learning (FSL) has attracted growing attention recent year.
It targets to help machines achieve or even surpass human beings' level with scarce labeled samples.
Generally, in FSL-based classification tasks, the current popular model usually includes two components: 
\textbf{(1)} Pre-train. Employ the base data $\mathcal{D}_{base}$ to generate a convolutional neural network (CNN) based feature extraction model (FEM). 
\textbf{(2)} Meta-test. First, extract the features of novel data $\mathcal{D}_{novel} = \{\mathcal{S}, \mathcal{U}, \mathcal{Q}\}$, where $\mathcal{S}$, $\mathcal{U}$ and $\mathcal{Q}$ denote support set, unlabeled set and query set. Next, design a classifier to recognize the query samples $\mathcal{Q}$.
Notably, $\mathcal{D}_{novel}$ has totally different categories from $\mathcal{D}_{base}$. For more details, please refer to \textbf{Section \ref{sec: Problem Formulation}}.


\begin{figure*}
	\begin{center}
		\includegraphics[width=0.9\linewidth]{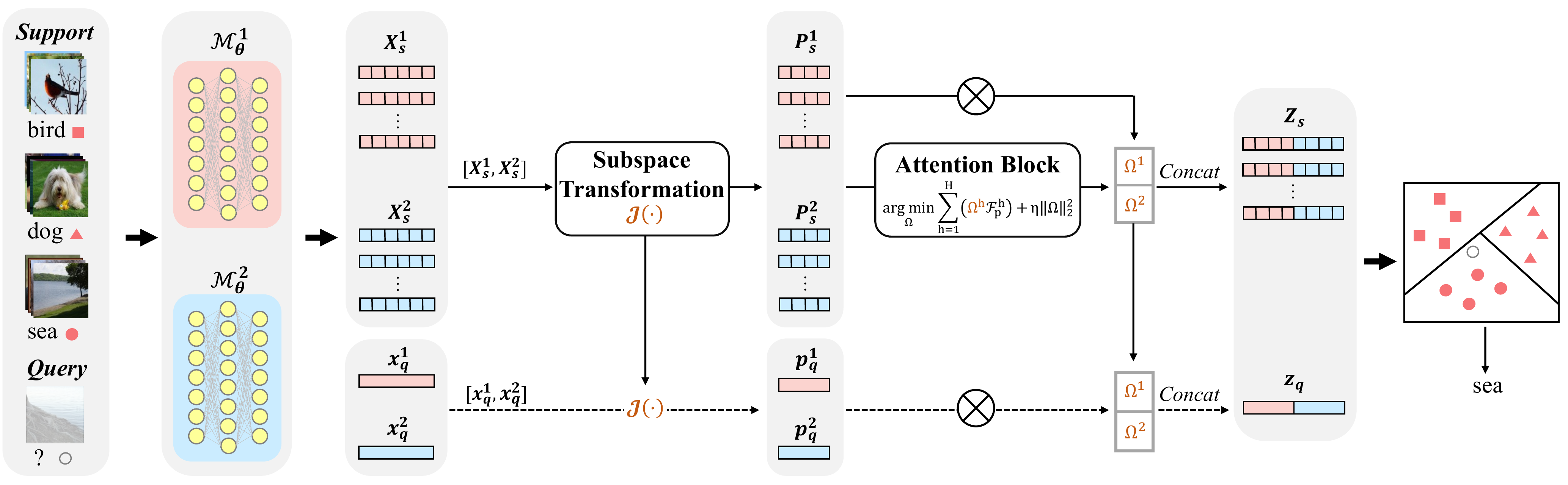}
	\end{center}
	\caption{
The framework of MHFC on inductive few-shot learning (IFSL). 
	Assume we have two heads of feature extraction models (FEMs), e.g., $\mathcal{M}_{\theta}^h$, where $h=[1,2]$ denotes the $h_{th}$ head. Different colors correspond to various features. There are a total of $5$ steps. 
	\textbf{(1)} Input images to FEMs and obtain the support features $\mathbf{X}_s^h$ and query feature $\mathbf{x}_q^h$.
	\textbf{(2)} Transform multi-head features to a unified space to obtain the novel features $\mathbf{P}_s^h$, $\mathbf{p}_q^h$.
	\textbf{(3)} Employ the transformed multi-head support features (e.g., $\mathbf{P}_s^h$) to learn the combination weights $\mathbf{\Omega}$ for each head, and assign them to the corresponding heads of features.
	\textbf{(4)} Collaboratively represent the samples by concatenating the weighted multi-head features.
	\textbf{(5)} Exploit the collaborative support features to construct classifier and recognize query sample.}
	\label{fig: Flowcharts}
\end{figure*}
    
In the pre-train phase, researchers have designed a variety of classical FEMs for base samples.
Each kind of model has a distinct emphasis. As examples, several may focus more attention on the contour information, whereas others may lay particular emphasis on the texture information.
Thus, the description based on one category of features is often one-sided for samples, not sufficiently accurate.
Besides, due to the negative influence of cross-domain 
(e.g., $\mathcal{D}_{base} \rightarrow \mathcal{D}_{novel}$, 
the trained FEM can not adapt to the novel class flawlessly), the novel data distribution may have a certain degree of deviation compared with the ground truth distribution. We dub this fundamental problem as a distribution-shift-problem (DSP).

Recent efforts on reducing the impact of DSP are generally based on designing a more robust and adaptive FEM to generate better features, such as introducing meta-learning strategy \cite{finn2017model} \cite{bertinetto2019metalearning};
self-supervised learning strategy \cite{mangla2020charting} \cite{rodriguez2020embedding}; knowledge distillation strategy \cite{tian2020rethinking} \cite{rizve2021exploring}. Such methods merely weaken the influence of the DSP to a certain extent. 
This paper proposes to tackle the issue from a multi-modal learning perspective.
We denote the multiple features extracted from different FEMs as multi-head features.
Since the single-head feature is only one-sided for the sample, why do not attempt to conduct multi-head features fusion? We illustrate a t-SNE \cite{van2008visualizing} visualization of different kinds of features in \textbf{Figure \ref{fig: tsne}}. 

However, two challenges prevent the idea.
\textbf{(1)} Since the independent FEMs, the extracted multi-head features are in separate spaces, which exists a problem of inconsistent measurement scales. 
Thus, the first challenge is how to align the multi-head features.
\textbf{(2)} 
Different features are suitable for various tasks. For example, features focusing on the contour information should be more crucial than features concentrating on texture information for classification tasks. 
Therefore, the second challenge is how to weight multi-head features reasonably.

To tackle the DSP and overcome these two challenges, we propose a novel \textbf{Multi-Head Feature Collaboration (MHFC)} algorithm to improve the discrimination of sample's feature.  
MHFC is a simple non-parametric model that can directly fuse multi-head features extracted from the existing FEMs, such as ICI-Net~\cite{wang2020instance}, MetaOpt-Net~\cite{bertinetto2019metalearning}.
To solve the first challenge, we introduce a subspace learning method to transform the raw multi-head features to a unified space with reconstructed low-dimensional representation. 
It is also helpful to reduce redundant information.
Next, we design an attention block to automatically update combination weights for each head of the feature to solve the second challenge efficiently.
Finally, we generate the collaborative representation of the sample by concatenating the processed multi-head features. 
We illustrate the flowchart in \textbf{Figure \ref{fig: Flowcharts}}.

Besides, according to the data adopted in the design of the classifier, researchers categorize the FSL-based approaches as three sorts: \textbf{(1)} Inductive few-shot learning (IFSL). \textbf{(2)} Transductive few-shot learning (TFSL). \textbf{(3)} Semi-supervised few-shot learning (SSFSL). This paper extends the proposed MHFC to these three settings.

In summary, the main contributions focus on:

\textbf{(1)} 
We propose a novel method for FSL, dubbed as \textbf{Multi-Head Feature Collaboration (MHFC)} and extend it to three FSL-based settings. It addresses the DSP by comprehensively considering the multi-head features to improve the discrimination of sample's representation.

\textbf{(2)}
Compared to tuning the network to solve DSP, the proposed MHFC is more straightforward and effective. It is a simple non-parametric method that directly fuses multi-head features extracted from the existing FEMs.
Besides, the more robust feature representation enables the proposed MHFC to achieve amazing performance especially when the samples are extremely scarce (e.g., inductive setting on $5$-way $1$-shot case, please refer to \textbf{Table \ref{table: comparison_results_mini_tiered}, \ref{table: comparison_results_cifar_fc100}}). Thus, we consider this paper may be meaningful to be applied in reality.

\textbf{(3)}
We evaluate the proposed method on four benchmark datasets (mini-ImageNet, tiered-ImageNet, CIFAR-FS, FC100) and achieve significant improvements of \textbf{2.1\%}-\textbf{7.8\%} compared with other state-of-the-art methods. Besides, to prove that the proposed method can overcome DSP, we design the cross-domain experiments (e.g., mini-ImageNet $\rightarrow$ CUB) and achieve far better performance than state-of-the-art methods of at least \textbf{7.8\%}.

\section{Related Work}
\label{sec: Related Work}
\subsection{Few-Shot Learning}
In the past decade, FSL based works have attracted lots of attention. Researchers have proposed various classical frameworks to solve this problem. We list the two most popular types, including 
\textbf{(1)} Meta-learning based methods, such as MAML \cite{finn2017model}, Reptile \cite{nichol2018first}, LEO \cite{rusu2019meta}, which purpose to obtain a universal model to rapidly adapt to new tasks. 
\textbf{(2)} Metric learning based methods, focusing on looking for ideal distance metrics to strengthen model's robustness, including ProtoNet \cite{snell2017prototypical}, MetaOpt \cite{bertinetto2019metalearning}, TADAM \cite{oreshkin2018tadam} $et$ $al.$
In addition, all these methods can be split into another taxonomy, e.g. inductive few-shot learning (IFSL), transductive few-shot learning (TFSL), and semi-supervised few-shot learning (SSFSL). For example, MAML \cite{finn2017model}, LEO \cite{rusu2019meta}, S2M2 \cite{mangla2020charting} $et$ $al.$ are based on inductive setting; DPGN \cite{yang2020dpgn}, TEAM \cite{qiao2019transductive}, SIB \cite{hu2020empirical} $et$ $al.$ are TFSL methods; and LST \cite{li2019learning}, EPNet \cite{rodriguez2020embedding}, ICI \cite{wang2020instance} $et$ $al.$ are based on semi-supervised setting. 

\subsection{Multi-Modal Learning}
Just as every coin has two sides, it would be incomplete to define objects from a single perspective. Therefore, multi-modal learning has received wide attention in recent years.
There exist lots of classical methods and corresponding applications. For example, 
Liu $et$ $al.$ proposed a sparse coding based multi-modal method MHDSC \cite{liu2014multiview} for image annotation task; 
Liu $et$ $al.$ proposed SPM-CRC \cite{liu2019weighted}, which improves the collaborative representation model from multi-modal learning to classify remote sensing images; 
Jan~$et$ $al.$ proposed MVCCA~\cite{rupnik2010multi} and employed it in natural language processing.
Liu $et$ $al.$ proposed MHL \cite{liu2017multi} to solve Alzheimer’s Disease Predicting problem; 
Zhang~$et$ $al.$ proposed IMHL~\cite{zhang2018inductive}, which is an inductive hypergraph learning from multi-modal and applied it for $3D$ object recognition.
All these methods may help FSL, and some multi-model based FSL methods were proposed, including \cite{dvornik2019diversity} (introduced ensemble strategy in pretrain), \cite{yue2020interventional} (adopted multi-head classifier in fine-tuning ), \cite{dvornik2020selecting} (fused multi-domain representation). This paper makes an orthogonal contribution towards efficient semantic-preserving pretraining.


\subsection{Subspace Learning}
Subspace learning is capable of transferring the samples to another representation. It is an efficient dimension reduction method. Here, we list several classical subspace learning methods, which have been employed in the proposed MHFC. The first one is Principal Component Analysis (PCA) \cite{tipping1999probabilistic}. It projects the raw features to a low-dimensional space by using singular value decomposition.
The second one is Locally Linear Embedding (LLE) \cite{roweis2000nonlinear}, which tries to preserve distances within local according to seeking a low-dimensional projection of the data. The last one is Laplacian Eigenmap (LE) \cite{belkin2002laplacian}. It exploits a spectral decomposition of the graph Laplacian to project the raw data to a low-dimensional representation.
All these methods are helpful for our MHFC, and we will show the comparison results in \textbf{Section\ref{subsubsec: Multi-Head Feature Transformation}}.

\section{Problem Formulation}
\label{sec: Problem Formulation}
In this section, we introduce the complete procedure in details. It is composed of two phases, including pre-train and meta-test.
i) In pre-train phase, given base data $\mathcal{D}_{base} = \{(x_i,y_i) |{\kern 1pt} y_i \in \mathcal{C}_{base} \}_{i=1}^{N_{base}}$, where $x$ and $y$ denote the sample and corresponding label, respectively. $N_{base}$ denotes the total number of base data.
$\mathcal{C}_{base}$ indicates the base category set. We train the CNN-based FEM $\mathcal{M}_{\theta}(\cdot)$ on $\mathcal{D}_{base}$, where $\theta$ indicates the parameters in CNN. 
In this paper, divers FEMs are employed to extract features for different heads, and we define the FEM on the $h_{th}$ head as $\mathcal{M}_{\theta}^h(\cdot)$, where $h=1,2,\cdots,H$, more details please refer to \textbf{Section \ref{subsec: Multi-Head Feature Extraction Model}}.

ii) For the meta-test phase, the novel data $\mathcal{D}_{novel} = \{(x_j,y_j) |{\kern 1pt} y_j \in \mathcal{C}_{novel} \}_{j=1}^{N_{novel}}$, where $\mathcal{C}_{novel}$ denotes the novel category set, $N_{novel}$ denotes the number of novel data. $C_{base} {\kern 2pt} \cap {\kern 2pt} C_{novel} = \emptyset$. $\mathcal{D}_{novel}$ includes three components, e.g., $\mathcal{D}_{novel} = \{\mathcal{S}, \mathcal{U}, \mathcal{Q}\}$, where $\mathcal{S}$, $\mathcal{U}$ and $\mathcal{Q}$ denote support set, unlabeled set and query set, 
$\mathcal{S} {\kern 2pt} \cap {\kern 2pt} \mathcal{U}  = \emptyset$, 
$\mathcal{S} {\kern 2pt} \cap {\kern 2pt} \mathcal{Q}  = \emptyset$,
$\mathcal{Q} {\kern 2pt} \cap {\kern 2pt} \mathcal{U}  = \emptyset$.
In this phase, we first utilize the trained FEM to extract $\mathcal{D}_{novel}$'s feature, then design a classifier to classify $\mathcal{Q}$. There exist three settings to design the classifier, e.g., inductive setting, transductive setting, and semi-supervised setting. \textbf{Section \ref{subsec: MHFC for Few-Shot Learning}} show more details.
We follow standard $C$-$way$-$T$-$shot$ per episode as \cite{wang2020instance} for classification, where $C$-$way$ denotes $C$ classes, and $T$-$shot$ indicates $T$ samples per class. We average the accuracies of all the episodes with $95\%$ confidence intervals as the final result.

\section{Methodology}
\label{sec: Methodology}
In this section, we first briefly review the linear regression classifier as an example to introduce our method. Then, we propose the novel \textbf{Multi-Head Feature Collaboration (MHFC)} algorithm to collaboratively represent the samples with processed multi-head features.
Next, we extend the proposed MHFC to three kinds of few-shot learning settings.
Finally, we introduce the employed multi-head feature extraction model.


\subsection{Review of Linear Regression Classifier}
\label{subsec: Review of  Linear Regression Classifier}
This proposed method pays attention to the various features that can integrate all types of conventional classification strategy (such as linear regression, support vector machine, logistic regression). 
In this paper, we employ a simple regularized linear regression model as an example to show the details of our method. We formulate the objective function as:
\begin{equation}
\begin{split}
        \mathop {\arg \min}\limits_{\mathbf{W}} \mathcal{F}
        = \left\| \mathbf{Y} - \mathbf{W} \mathbf{X} \right\|_F^2
        + \mu \left \| \mathbf{W} \right\|_F^2
\end{split}
\label{eqa: linear_regression_obj}
\end{equation}
where $\left\| \cdot \right\|_F$ represents $\left( \cdot \right)$'s Frobenius-norm.
$\mu$ is the hyperparameter.
$\mathbf{X}=[{\mathbf{x}}_1,{\mathbf{x}}_2,\dots,{\mathbf{x}}_{N}] \in 
\mathbb{R} ^{dim1 \times N}$,

$\mathbf{Y}=[{\mathbf{y}}_1,{\mathbf{y}}_2,\dots,{\mathbf{y}}_{N}] \in \mathbb{R} ^{C \times N}$, $dim1$ and $N$ indicate the dimension and number of labeled samples, respectively. $C$ denotes the number of categories.
${\mathbf{x}}_n,{\mathbf{y}}_n$ ($n = 1, 2, \dots, N$) denote the feature embedding vector and one-hot label vector of the $n_{th}$ sample.
$\mathbf{W} \in \mathbb{R} ^{C \times dim1}$ represents the to-be-learned classifier. 
We directly optimize the objective function and obtain the $\mathbf{W}$ as:
\begin{equation}
\begin{split}
        \mathbf{W} = \mathbf{Y}{\mathbf{X}}^T \left(\mathbf{X}{\mathbf{X}}^T+\mu \mathbf{I} \right)^{-1}
\end{split}
\label{eqa: LR_classifier}
\end{equation}
where $\mathbf{I}$ is the identity matrix.
Following, given a testing sample embedding $\mathbf{x}_{ts} \in \mathbb{R}^{dim1}$, we predict the $\mathbf{x}_{ts}$'s category by:
\begin{equation}
\begin{split}
        \mathcal{A}(\mathbf{x}_{ts}) = max
        \left\{
        \mathbf{W} \mathbf{x}_{ts}  
        \right\}
\end{split}
\label{eqa: single_view_predict}
\end{equation}
where $max$ denotes an operator to obtain the index of the max value in the vector.

\subsection{Multi-Head Feature Collaboration}
\label{subsec: Multi-Head Feature Collaboration}

\subsubsection{\textbf{Multi-Head Feature Transformation}}
\label{subsubsec: Multi-Head Feature Transformation}
Since one kind of incomplete feature cannot reflect a sample well, we introduce various features to represent samples from different heads collaboratively. Assume that we have $H$ heads in total.
Each head corresponds to one kind of feature $\mathbf{X}^h$, where $h=1,2,\cdots,H$.
As mentioned in \textbf{Section \ref{sec: introduction}}, the multi-head features are in separate spaces, which exists a problem of inconsistent measurement scales.

To address this challenge, we introduce a conventional subspace learning method (denoted as $\mathcal{J}(\cdot)$) to transform the raw features into a unified space with reconstructed low-dimensional representation. 
Specifically, we treat $H$ heads of the same sample as $H$ samples, and denote the features of the expanded dataset as $\mathbf{X}_{exp} =[\mathbf{X}^1,\mathbf{X}^2,\cdots,\mathbf{X}^H] \in \mathbb{R}^{dim1 \times (N \times H)}$. We conduct subspace learning operation $\mathcal{J}(\mathbf{X}_{exp})$ and obtain the novel features $\mathbf{X}_{exp}' = [\mathbf{P}^1,\mathbf{P}^2,\cdots,\mathbf{P}^H] \in \mathbb{R}^{dim2 \times (N \times H)}$, where $\mathbf{P}^h \in \mathbb{R}^{dim2 \times N}$ denotes the feature on the $h_{th}$ head after subspace transformation. $dim2$ denotes the novel dimension.

\subsubsection{\textbf{Multi-Head Feature Attention Block}}
\label{subsubsec: Multi-Head Feature Attention Block}
Consider that the significance of these features is different per episode.
We try to find the optimal combination weights $\mathbf{\Omega} = [\Omega^1,\Omega^2,\dots,\Omega^H]^T$ to let these features have different influences for the final decision, where $\mathbf{\Omega}$ is a weight vector, $\Omega^h (h=1,2,\cdots,H)$ denotes the $h_{th}$ element in $\mathbf{\Omega}$.
We first employ the transformed feature $\mathbf{P}^h$ to replace $\mathbf{X}^h$ and obtain a novel classifier $\mathbf{W}_p^h \in \mathbb{R}^{C \times dim2}$ 
according to \textbf{Equation \ref{eqa: LR_classifier}}, where $\mathbf{W}_p^h$ indicates the trained novel classifier on the $h_{th}$ head, which can be formulated as:
\begin{equation}
\begin{split}
        \mathbf{W}_p^h = \mathbf{Y}{\mathbf{P}^h}^T \left(\mathbf{P}^h{\mathbf{P}^h}^T+\mu \mathbf{I} \right)^{-1}
\end{split}
\label{eqa: LR_classifier_PCA}
\end{equation}

Then use $\mathbf{P}^h$ and $\mathbf{W}_p^h$ to re-calculate the objective function's loss on the $h_{th}$ head $\mathcal{F}_p^h$ by \textbf{Equation \ref{eqa: linear_regression_obj}}, which can be formulated as:
\begin{equation}
\begin{split}
        \mathcal{F}_p^h
        = \left\| \mathbf{Y} - \mathbf{W}_p^h \mathbf{P}^h \right\|_F^2
        + \mu \left \| \mathbf{W}_p^h \right\|_F^2
\end{split}
\label{eqa: linear_regression_obj_PCA}
\end{equation}

Next, exploit the $\mathcal{F}_p^h$ to compute the combination weights. The objective function can be formulated as:
\begin{equation}
\begin{split}
        \mathop {\arg \min}\limits_{\mathbf{\Omega}} 
        & = \sum_{h=1}^H 
        \left (
        \Omega^h \mathcal{F}_p^h 
        \right )
        + \eta \left\| \mathbf{\Omega} \right\|_2^2\\
		\text{s.t.} {\kern 4pt}  & \sum_{h=1}^H \Omega^h = 1 {\kern 4pt} \Omega^h \ge 0
\end{split}
\label{eqa: obj_compute_Omega}
\end{equation}
where 
$\Omega^h$ indicates the weight of $h_{th}$ head.
$\left\| \cdot \right\|_2$ represents $\left( \cdot \right)$'s $\ell_2$-norm.
$\eta$ is the parameter.
We introduce the Lagrangian to solve the problem, the \textbf{Equation \ref{eqa: obj_compute_Omega}} can be rewritten as:
\begin{equation}
\begin{split}
        \mathop {\arg \min}\limits_{\mathbf{\Omega, \zeta, \Lambda}} 
        \mathcal{G}=
        \sum_{h=1}^H 
        \left (
        \Omega^h \mathcal{F}_p^h 
        \right )
        + \eta \left\| \mathbf{\Omega} \right\|_2^2 
         - \zeta \left(\sum_{h=1}^H \Omega^h - 1 \right)
        -\mathbf{\Lambda}^T \mathbf{\Omega}
\end{split}
\label{eqa: obj_compute_Omega_Lagrangian}
\end{equation}
where $\zeta$ is a constant, $\mathbf{\Lambda} = [\Lambda^1,\Lambda^2,\dots,\Lambda^H]^T$ is a vector.
Assume $\hat{\mathbf{\Omega}}$, $\hat{\zeta}$, $\hat{\mathbf{\Lambda}}$ are the optimal solutions, we solved this problem as:
\begin{equation}
\begin{split}
        \hat{\Omega}^h = \frac{1}{2\eta} 
        max\left\{
        \frac{\sum_{h=1}^H \mathcal{F}_p^h}{H} + \frac{2\eta}{H} - \mathcal{F}_p^h - \hat{\Lambda}_{avg} ,0\right\}\\
\end{split}
\label{eqa: omega_final}
\end{equation}
where $\hat{\Lambda}_{avg}$ is a constant, denotes the average of $\hat{\mathbf{\Lambda}}$.
For the detailed optimization process, please refer to \textbf{Appendix A.2}.

\subsubsection{\textbf{Multi-Head Feature Collaborative Classifier}}
\label{subsubsec: Multi-Head Feature Collaborative Classifier}
After giving muli-head features the combination weights, we obtain the final collaborative feature $\mathbf{Z} =[\mathbf{z}_1,\mathbf{z}_2,\cdots,\mathbf{z}_N] \in \mathbb{R}^{dim3 \times N}$ by:
\begin{equation}
\begin{split}
        \mathbf{Z}
        \leftarrow
        \mathbf{z}_n = Concat 
        \left(
        \hat{\Omega}^1 {\kern 1pt} \mathbf{p}_n^1  ,
        \hat{\Omega}^2 {\kern 1pt} \mathbf{p}_n^2  ,
        \cdots,
        \hat{\Omega}^H {\kern 1pt} \mathbf{p}_n^H  
        \right)
\end{split}
\label{eqa: fea_emb_final}
\end{equation}
where $\mathbf{p}_n^h, \mathbf{z}_n^h (n=1,2,\cdots,N)$ denote the $n_{th}$ vector of $\mathbf{P}^h$ and $\mathbf{Z}^h$.

Next, according to \textbf{Equation \ref{eqa: LR_classifier}}, we use $\mathbf{Z}$ to replace $\mathbf{X}$ and obtain the final collaborative classifier $\mathbf{W}_z \in \mathbb{R}^{C \times dim3}$, which can be formulated as:
\begin{equation}
\begin{split}
        \mathbf{W}_z = \mathbf{Y}{\mathbf{Z}}^T \left(\mathbf{Z}{\mathbf{Z}}^T+\mu \mathbf{I} \right)^{-1}
\end{split}
\label{eqa: LR_classifier_final}
\end{equation}

Finally, given a testing sample feature $\mathbf{x}_{ts}^h, (h=1,2,\cdots,H)$, we first obtain the collaborative feature $\mathbf{z}_{ts} \in \mathbb{R}^{dim3}$ 
by \textbf{Equation \textbf{\ref{eqa: omega_final}, \ref{eqa: fea_emb_final}}}, 
then predict the label by: 
\begin{equation}
\begin{split}
        \mathcal{A}(\mathbf{z}_{ts}) = max
        \left\{
        \mathbf{W}_z \mathbf{z}_{ts}  
        \right\}
\end{split}
\label{eqa: final_predict}
\end{equation}

\subsection{MHFC for Few-Shot Learning}
\label{subsec: MHFC for Few-Shot Learning}
Define the feature of $\mathcal{D}_{novel}$ on the $h_{th}$ head as ${\mathbf{X}_{novel}^h}=[{\mathbf{X}_s^h}, {\mathbf{X}_u^h}, \\{\mathbf{X}_q^h}]$, where ${\mathbf{X}_s^h}=\mathcal{M}_{\theta}^h(\mathcal{S})$, ${\mathbf{X}_u^h}=\mathcal{M}_{\theta}^h(\mathcal{U})$, and ${\mathbf{X}_q^h}=\mathcal{M}_{\theta}^h(\mathcal{Q})$ denote the features of support, unlabeled, and query data on the $h_{th}$ head.
Thus, the $\left(\mathbf{X}_{novel}^h\right)$'s feature after transformation can be defined as $\mathbf{P}_{novel}^h=[{\mathbf{P}_s^h}, {\mathbf{P}_u^h}, {\mathbf{P}_q^h}]$.
Researchers employ different data to design the classifier, and these methods can be split into three settings, e.g., inductive setting, semi-supervised setting, and transductive setting.

\subsubsection{\textbf{Semi-Supervised MHFC}}
\label{subsubsec: Semi-Supervised MHFC}
Semi-supervised setting in few-shot learning adopts the support set $\mathcal{S}$ and unlabeled set $\mathcal{U}$ to train the classifier, and then predict the query label. In this paper, we extend our MHFC to the semi-supervised setting by introducing a simple self-training strategy. It can employ the unlabeled data to strengthen the classifier. We show the detailed steps as:

\textbf{(1)} 
Exploit the support data to train the basic classifier on each head, \textbf{Equation \ref{eqa: LR_classifier_PCA}} can be rewritten as:
\begin{equation}
\begin{split}
        \mathbf{W}_p^h = \mathbf{Y}_s{\mathbf{P}_s^h}^T \left(\mathbf{P}_s^h{\mathbf{P}_s^h}^T+\mu \mathbf{I} \right)^{-1}
\end{split}
\label{eqa: semi_classifier}
\end{equation}
where $\mathbf{Y}_s$ denotes the one-hot label matrix of support data.

\textbf{(2)} 
Calculate the combination weight on each head by:
\begin{equation}
\begin{split}
        \left\{\begin{array}{lllllll}
            \mathcal{F}_p^h
            = \left\| \mathbf{Y}_s - \mathbf{W}_p^h \mathbf{P}_s^h \right\|_F^2
            + \mu \left \| \mathbf{W}_p^h \right\|_F^2\\
            \hat{\Omega}^h = \frac{1}{2\eta} 
            max\left\{
            \frac{\sum_{h=1}^H \mathcal{F}_p^h}{H} + \frac{2\eta}{H} - \mathcal{F}_p^h - \hat{\Lambda}_{avg} ,0\right\}\\
        \end{array}\right.
\end{split}
\label{eqa: semi_weight}
\end{equation}

\textbf{(3)} 
Obtain the collaborative feature of support set and classifier by:
\begin{equation}
\begin{split}
        \left\{\begin{array}{lllllll}
            \mathbf{Z}_s
            \leftarrow
            \mathbf{z}_{sn} = Concat 
            \left(
            \hat{\Omega}^1 {\kern 1pt} \mathbf{p}_{sn}^1  ,
            \hat{\Omega}^2 {\kern 1pt} \mathbf{p}_{sn}^2  ,
            \cdots,
            \hat{\Omega}^H {\kern 1pt} \mathbf{p}_{sn}^H  
            \right)\\
            \mathbf{W}_z = \mathbf{Y}_s{\mathbf{Z}_s}^T \left(\mathbf{Z}_s{\mathbf{Z}_s}^T+\mu \mathbf{I} \right)^{-1}\\  
        \end{array}\right.
\end{split}
\label{eqa: semi_classifier_collaborative}
\end{equation}
where $\mathbf{p}_{sn}^h (n=1,2,\cdots,N)$ indicates the $n_{th}$ vector in $\mathbf{P}_{s}^h$.
$\mathbf{Z}_s$ represents the collaborative feature of support set.
$\mathbf{z}_{sn}$ denotes the $n_{th}$ vector in $\mathbf{Z}_{s}$.

  
    
    

    
     
    
     


\textbf{(4)} 
Utilize the trained classifier $\mathbf{W}_z$ to predict the unlabeled data $\mathcal{U}$ by:
\begin{equation}
\begin{split}
        \left\{\begin{array}{lllllll}
            \mathbf{Z}_u
            \leftarrow
            \mathbf{z}_{un} = Concat 
            \left(
            \hat{\Omega}^1 {\kern 1pt} \mathbf{p}_{un}^1,
            \hat{\Omega}^2 {\kern 1pt} \mathbf{p}_{un}^2,
            \cdots,
            \hat{\Omega}^H {\kern 1pt} \mathbf{p}_{un}^H  
            \right)\\            
            \mathbf{Y}_{pseudo} = 
            \mathbf{W}_z \mathbf{Z}_{u}  
            \\  
        \end{array}\right.
\end{split}
\label{eqa: semi_pseudo_label}
\end{equation}
where $\mathbf{p}_{un}^h (n=1,2,\cdots,N)$ indicates the $n_{th}$ vectors in $\mathbf{P}_{u}^h$.
$\mathbf{Z}_u$ represents the collaborative feature embedding of the unlabeled set.
$\mathbf{z}_{un}$ denotes the $n_{th}$ vector in $\mathbf{Z}_{u}$. 
$\mathbf{Y}_{pseudo}$ represents the predicted soft-pseudo-label of unlabeled data.

\textbf{(5)}
Select one most confidence sample through the ${\mathbf{Y}_{pseudo}}$ without putting back, the corresponding one-hot-pseudo-label and feature (after transformation) are defined as ${\mathbf{y}_{select}}$ and ${\mathbf{p}^h_{select}}$. Then, expand it to the support set by:
\begin{equation}
\begin{split}
        \left\{\begin{array}{ll}
            {\mathbf{P}^h_s} 
            = \left[{\mathbf{P}^h_s},{\mathbf{p}^h_{select}}  \right]\\
            {\mathbf{Y}_s} 
            = \left[{\mathbf{Y}_s},{\mathbf{y}_{select}}  \right]
        \end{array}\right.
\end{split}
\label{eqa: select_most_confidence_unlabel_data}
\end{equation}

\textbf{(6)}
Repeat \textbf{(1)}, \textbf{(2)} \textbf{(3)}, \textbf{(4)}, \textbf{(5)} until the performance of classifier is stable. 

\textbf{(7)}
Obtain the collaborative feature embedding of query data by:
\begin{equation}
\begin{split}
            \mathbf{Z}_q
            \leftarrow
            \mathbf{z}_{qn} = Concat 
            \left(
            \hat{\Omega}^1 {\kern 1pt} \mathbf{p}_{qn}^1  ,
            \hat{\Omega}^2 {\kern 1pt} \mathbf{p}_{qn}^2  ,
            \cdots,
            \hat{\Omega}^H {\kern 1pt} \mathbf{p}_{qn}^H  
            \right)\\
\end{split}
\label{eqa: semi_query_collaborative_fea}
\end{equation}
where $\mathbf{p}_{qn}^h (n=1,2,\cdots,N)$ indicates the $n_{th}$ vector in $\mathbf{P}_{q}^h$.
$\mathbf{Z}_q$ represents the collaborative feature of query data.
$\mathbf{z}_{qn}$ denotes the $n_{th}$ vector in $\mathbf{Z}_{q}$.

\textbf{(8)}
Finally, employ the optimal classifier to predict the query label by:
\begin{equation}
\begin{split}
        \mathcal{A}(\mathbf{Z}_{q}) = max
        \left\{
        \mathbf{W}_z \mathbf{Z}_{q}  
        \right\}
\end{split}
\label{eqa: predict_query}
\end{equation}

We summarize the Algorithm in \textbf{Appendix A.1}.

\subsubsection{\textbf{Inductive MHFC}}
\label{subsubsec: Inductive MHFC}
Unlike semi-supervised few-shot learning (SSFSL), inductive few-shot learning (IFSL) based methods only employ the support data to train the classifier and predict the query's category.
IFSL can be viewed as the special case of SSFSL (e.g., there not exist unlabeled data).
Thus, we can use the steps (e.g., \textbf{(1)}, \textbf{(2)}, \textbf{(3)}, \textbf{(7)}, \textbf{(8)}) to implement the inductive MHFC.

\subsubsection{\textbf{Transductive MHFC}}
\label{subsubsec: Transductive MHFC}
In transductive few-shot learning (TFSL), besides the support data's features and label information, researchers also apply the features of query data to construct the classifier and then predict the query label.
To implement the tansductive MHFC, we need to make some adjustments to the step \textbf{(4)} in \textbf{Section \ref{subsubsec: Semi-Supervised MHFC}}. We re-represent \textbf{(4)} as:

\textbf{(9)} 
Utilize the trained classifier $\mathbf{W}_z$ to predict the query data $\mathcal{Q}$ by:
\begin{equation}
\begin{split}
        \left\{\begin{array}{lllllll}
            \mathbf{Z}_q
            \leftarrow
            \mathbf{z}_{qn} = Concat 
            \left(
            \hat{\Omega}^1 {\kern 1pt} \mathbf{p}_{qn}^1,
            \hat{\Omega}^2 {\kern 1pt} \mathbf{p}_{qn}^2,
            \cdots,
            \hat{\Omega}^H {\kern 1pt} \mathbf{p}_{qn}^H  
            \right)\\            
            \mathbf{Y}_{pseudo} = 
            \mathbf{W}_z \mathbf{Z}_{q}  
            \\  
        \end{array}\right.
\end{split}
\label{eqa: trans_pseudo_label}
\end{equation}
where $\mathbf{p}_{qn}^h (n=1,2,\cdots,N)$ indicates the $n_{th}$ vectors in $\mathbf{P}_{q}^h$.
$\mathbf{Z}_q$ represents the collaborative feature embedding of the unlabeled set.
$\mathbf{z}_{qn}$ denotes the $n_{th}$ vector in $\mathbf{Z}_{q}$. $\mathbf{Y}_{pseudo}$ represents the predicted soft-pseudo-label of query data.

After that, we achieve transductive MHFC by steps \textbf{(1)}, \textbf{(2)}, \textbf{(3)},  \textbf{(9)}, \textbf{(5)}, \textbf{(6)}, \textbf{(8)}.

\subsection{Multi-Head Feature Extraction Model}
\label{subsec: Multi-Head Feature Extraction Model}


The multi-head features we adopted come from different feature extraction models (FEMs).
As examples:
\textbf{(1)}
Standard feature (Std-Fea), the FEM utilizes a standard CNN-based classification structure, such as \cite{wang2020instance}.
\textbf{(2)}
Meta feature (Meta-Fea), the FEM introduces the meta-learning strategy to the network, just like \cite{bertinetto2019metalearning}.
\textbf{(3)}
Self-supervised-feature (SS-Fea), the FEM adds auxiliary losses to the standard CNN-based classification structure from a self-supervised perspective to strengthen the robustness of the network, similar as \cite{mangla2020charting}.
We discuss the results of all kinds of stacking ways in \textbf{Appendix B.2}.


In this paper, we merely fuse two kinds of SS-Feas for most of the experiments as an example for convenience. 
For the first category, we design the FEM by introducing standard classification loss $\mathcal{L}_c$ and auxiliary rotation loss $\mathcal{L}_r$. $\mathcal{L}_c$ can be formulated as:
\begin{equation}
\begin{split}
        \mathcal{L}_{c}
        =-\sum_{c}
         y_{(c,x)}log(p_{(c,x)})\\
\end{split}
\label{eqa: classification_loss}
\end{equation}
where $c \in \mathcal{C}_{base}$ denotes the $c_{th}$ class. $y_{(c,x)}$, $p_{(c,x)}$ indicate the probabilities that the truth label and predicted label of $x_{th}$ sample belongs to $c_{th}$ class.
Then, we rotate each sample to $r$ degree and $r \in \mathcal{C_R} = \{0^{\circ}, 90^{\circ}, 180^{\circ}, 270^{\circ} \}$. We define rotation loss as:
\begin{equation}
\begin{split}
        \mathcal{L}_{r}
        =-\sum_{r}
         y_{(r,x)}log(p_{(r,x)})\\
\end{split}
\label{eqa: rotation_loss}
\end{equation}
where $y_{(r,x)}$, $p_{(r,x)}$ indicate the probabilities that the truth label and predicted label of $x_{th}$ sample belongs to $r_{th}$ class.
Thus, the first loss function is defined as $\mathcal{L}_c + \mathcal{L}_r$, and the feature based on this kind of FEM is dubbed as SS-R-Fea.

The second feature is denoted as SS-M-Fea, which extracted from another category of self-supervised FEM.
Specifically, this FEM adds the loss $\mathcal{L}_c$ and auxiliary mirror loss $\mathcal{L}_m$ to the neural network 
to predict image mirrors.
Assume that there are $m$ ways and $m \in \mathcal{C_M} = \{vertically, horizontally, diagonally \}$, we define the mirror loss as:
\begin{equation}
\begin{split}
        \mathcal{L}_{m}
        =-\sum_{m}
         y_{(m,x)}log(p_{(m,x)})\\
\end{split}
\label{eqa: mirror_loss}
\end{equation}
where $y_{(m,x)}$, $p_{(m,x)}$ indicate the probabilities that the truth label and predicted label of $x_{th}$ sample belongs to $m_{th}$ class.
Next, we summarize the loss function as $\mathcal{L}_c + \mathcal{L}_m$.

\section{Experiments}
In this section, we first briefly review the benchmark datasets and show the implementation details. Then, we list the experimental results in \textbf{Table \ref{table: comparison_results_mini_tiered}, \ref{table: comparison_results_cifar_fc100}} and analyse them. Next, we perform ablation studies to discuss the factors that influence MHFC's performance, e.g., multi-head feature fusion, subspace transformation, and attention block that calculates the combination weights.
In the following, we take a cross-domain experiment to further evaluate the ability and robustness of the proposed method.
We conduct all the experiments on a Tesla-$V100$ GPU with $32G$ memory. All the source codes will be made available to the public.

\begin{table*}[!t]
\caption{The $5$-way few-shot classification accuracies on mini-ImageNet and tiered-ImageNet with $95\%$ confidence intervals over $600$ episodes. 
$(\cdot)^{\star}$, $(\cdot)^{\dagger}$, and $(\cdot)^{\ddagger}$ in Table \ref{table: comparison_results_mini_tiered}, \ref{table: comparison_results_cifar_fc100}, \ref{table: comparison_results_cross-domain} indicate inductive, transductive, and semi-supervised settings, respectively. 
$(\cdot)^{\star\star}$ in Table \ref{table: comparison_results_mini_tiered}, \ref{table: comparison_results_cifar_fc100} and \ref{tab: fix_weight} denotes the non-standardized inductive setting, which adopts the query feature when reducing the feature's dimension.
4CONV, ResNet12, ResNet18 and WRN are the exploited FEM's architectures. 
The $(80)$, $(100)$ in the semi-supervised setting indicate the number of employed unlabeled samples per class. The top two results are shown in red and blue, respectively.}
\begin{center}
\setlength{\tabcolsep}{1.5mm}{
\begin{tabular}{lcccccc}
\toprule 
\multicolumn{1}{l}{\multirow{2}{*}{\textbf{Method}}}
& \multicolumn{1}{c}{\multirow{2}{*}{\textbf{Backbone}}} 
& \multicolumn{2}{c}{\textbf{mini-ImageNet}} 
& \multicolumn{2}{c}{\textbf{tiered-ImageNet}} 
\\ 
\cmidrule(l){3-6}
\multicolumn{1}{c}{}
& \multicolumn{1}{c}{}                          
& \textbf{$5$-way $1$-shot} & \textbf{$5$-way $5$-shot}  & \textbf{$5$-way $1$-shot}  & \textbf{$5$-way $5$-shot} \\  
\midrule
\midrule
Baseline$^\star$ \cite{chen2019closer} (ICLR,2019) & ResNet18
& $51.75 \pm 0.80$   & $74.27 \pm 0.63$    & -   & -          \\
Baseline++$^\star$ \cite{chen2019closer} (ICLR,2019) & ResNet18
& $51.87 \pm 0.77$   & $75.68 \pm 0.63$    & -   & -          \\
TapNet$^\star$ \cite{yoon2019tapnet} (ICML,2019) & ResNet12
& $61.65 \pm 0.15$   & $76.36 \pm 0.10$    & $63.08 \pm 0.15$   & $80.26 \pm 0.12$          \\
LEO$^\star$ \cite{rusu2019meta} (ICLR,2019) & WRN
& $61.76 \pm 0.08$   & $77.59 \pm 0.12$    & $66.33 \pm 0.05$    & $81.44 \pm 0.09$      \\
AM3$^\star$ \cite{xing2019adaptive} (NIPS,2019) & ResNet12
& $65.30 \pm 0.49$   & $78.10 \pm 0.36$    & $69.08 \pm 0.47$    & $82.58 \pm 0.31$      \\
CTM$^\star$ \cite{li2019finding} (CVPR,2019) & ResNet18
& $64.12 \pm 0.82$   & $80.51 \pm 0.13$    & -                   & -          \\
MABAS$^\star$ \cite{kim2020model} (ECCV,2020) & ResNet12
& $64.21 \pm 0.82$    & $81.01 \pm 0.57$    & -   & -   \\
MELR$^\star$  \cite{fei2021melr} (ICLR,2021)  & ResNet12
& {\color{blue}$\textbf{67.40} \pm 0.43$}    & {\color{red}$\textbf{83.40} \pm 0.28$}    & {\color{blue}$\textbf{72.14} \pm 0.51$}   & {\color{blue}$\textbf{87.01} \pm 0.35$}    \\
\midrule
\textbf{Our MHFC}$^{\star\star}$ & ResNet12
& {\color{red} $\textbf{73.10} \pm 1.00$}   & {\color{blue}$\textbf{81.75} \pm 0.56$}    & {\color{red} $\textbf{82.10} \pm 1.03$}    & {\color{red}$\textbf{87.99} \pm 0.60$}    \\
\midrule
\midrule
TPN$^\dagger$ \cite{liu2019learning} (ICLR,2019) & 4CONV
& $55.51 \pm 0.86$    & $69.86 \pm 0.65$    & $59.91 \pm 0.94$   & $73.30 \pm 0.75$   \\
TEAM$^\dagger$ \cite{qiao2019transductive} (ICCV,2019) & ResNet12
& $60.07 \pm 0.63$   & $75.90 \pm 0.52$    & -    & -    \\
Fine-tuning$^\dagger$ \cite{dhillon2020baseline} (ICLR,2020) & WRN
& $65.73 \pm 0.68$    & $78.40 \pm 0.52$    & {\color{blue}$\textbf{73.34} \pm 0.71$}   & $85.50 \pm 0.50$   \\
DPGN$^\dagger$  \cite{yang2020dpgn} (CVPR,2020)  & ResNet12
& {\color{blue}$\textbf{67.77} \pm 0.32$}    & {\color{blue}$\textbf{84.60} \pm 0.43$}    & $72.45 \pm 0.51$   & {\color{blue}$\textbf{87.24} \pm 0.39$}   \\
ODE$^\dagger$  \cite{xu2021learning} (CVPR,2021)  & ResNet12
& $67.76 \pm 0.46$    & $82.71 \pm 0.31$    & $71.89 \pm 0.52$   & $85.96 \pm 0.35$ \\
\midrule
\textbf{Our MHFC}$^\dagger$ & ResNet12
& {\color{red} $\textbf{74.81} \pm 1.12$}   & {\color{red} $\textbf{85.58} \pm 0.61$}    & {\color{red} $\textbf{83.95} \pm 1.13$}   & {\color{red} $\textbf{90.75} \pm 0.58$}    \\
\midrule
\midrule
TPN$^\ddagger$ \cite{liu2019learning} (ICLR,2019) & 4CONV
& $52.78 \pm 0.27$    & $66.42 \pm 0.21$    & $55.74 \pm 0.29$   & $71.01 \pm 0.23$   \\
LST$^\ddagger$ (100) \cite{li2019learning} (NIPS,2019) & ResNet12
& $70.10 \pm 1.90$   & $78.70 \pm 0.80$    & $77.70 \pm 1.60$    & $85.20 \pm 0.80$     \\
EPNet$^\ddagger$ ($100$) \cite{rodriguez2020embedding} (ECCV,2020)  & ResNet12   
&  {\color{blue} $\textbf{75.36} \pm 1.01$}    &  {\color{blue} $\textbf{84.07} \pm 0.60$}    & $81.79 \pm 0.97$   & $88.45 \pm 0.61$   \\
TransMatch$^\ddagger$ ($100$) \cite{yu2020transmatch} (CVPR,2020) & WRN
& $63.02 \pm 1.07$    & $81.19 \pm 0.59$    & -          & -         \\
ICI$^\ddagger$ ($80$) \cite{wang2020instance} (CVPR,2020)  & ResNet12
& $71.41$             & $81.12$             &  {\color{blue} $\textbf{85.44}$}   &  {\color{blue} $\textbf{89.12} $}   \\
\midrule
\textbf{Our MHFC}$^\ddagger$ ($80$) & ResNet12
& $79.26 \pm 1.14$   & $87.30 \pm 0.55$    & {\color{red} $\textbf{87.57} \pm 1.03$}    & $ 91.80 \pm 0.56 $    \\
\textbf{Our MHFC}$^\ddagger$ ($100$) & ResNet12
& {\color{red} $\textbf{79.76} \pm 1.16$}  & {\color{red} $\textbf{87.64} \pm 0.53$}  & $87.56 \pm 1.04$    
& {\color{red} $\textbf{91.90} \pm 0.56$}   \\


\bottomrule
\end{tabular}
}
\label{table: comparison_results_mini_tiered} 
\end{center}
\end{table*}

\subsection{Datasets}
We carry out experiments on five benchmark datasets, including mini-ImageNet \cite{vinyals2016matching}, tiered-ImageNet \cite{ren2018meta}, CIFAR-FS \cite{bertinetto2019metalearning}, FC100 \cite{oreshkin2018tadam}, and CUB \cite{wah2011caltech}. 
Both \textbf{mini-ImageNet} and \textbf{tiered-ImageNet} are the subsets of ImageNet dataset \cite{russakovsky2015imagenet}. mini-ImageNet consists of $100$ classes and tiered-ImageNet contains $608$ classes. For both datasets, the number of images for each class is $600$ and the size of each image is $84 \times 84$. We follow standard split as \cite{wang2020instance}, selecting $64$ classes as the base set, $16$ classes as the validation set, $20$ classes as the novel set for mini-ImageNet, and selecting $351$ classes as the base set, $97$ classes as the validation set, $160$ classes as the novel set for tiered-ImageNet. 
Both \textbf{CIFAR-FS} and \textbf{FC100} are the subsets of CIFAR-100 dataset \cite{krizhevsky2009learning}, and consist of $100$ classes.
We follow the split introduced in \cite{bertinetto2019metalearning} to divide CIFAR-FS into $64$ classes as base set, $16$ classes as validation set, $20$ classes as novel set, and divide FC100 into $60$ classes as base set, $20$ classes as validation set, $20$ classes as novel set. All the image size is $32 \times 32$.
\textbf{CUB} totally includes $11,788$ images with $200$ categories. We follow the setting in ICI \cite{wang2020instance} to split it into $100$ classes as base set, $50$ classes as validation set and $50$ classes as novel set. The images are cropped into $84 \times 84$.

\begin{table*}[!t]
\caption{The $5$-way few-shot classification accuracies on CIFAR-FS and FC100 with $95\%$ confidence intervals over $600$ episodes. The top two results are shown in red and blue, respectively.}
\begin{center}
\setlength{\tabcolsep}{1.5mm}{
\begin{tabular}{lcccccc}
\toprule 
\multicolumn{1}{l}{\multirow{2}{*}{\textbf{Method}}}
& \multicolumn{1}{c}{\multirow{2}{*}{\textbf{Backbone}}} 
& \multicolumn{2}{c}{\textbf{CIFAR-FS}} 
& \multicolumn{2}{c}{\textbf{FC100}} 
\\ 
\cmidrule(l){3-6}
\multicolumn{1}{c}{}
& \multicolumn{1}{c}{}                          
& \textbf{$5$-way $1$-shot} & \textbf{$5$-way $5$-shot}  & \textbf{$5$-way $1$-shot}  & \textbf{$5$-way $5$-shot} \\ 

\midrule
\midrule
ProtoNet$^\star$ \cite{snell2017prototypical} (NIPS,2017) & 4CONV
& $55.50 \pm 0.70$   & $72.00 \pm 0.60$    & $35.30 \pm 0.60$    & $48.60 \pm 0.60$    \\
MAML$^\star$ \cite{finn2017model} (ICML,2018) & 4CONV
& $58.90 \pm 1.90$   & $71.50 \pm 1.00$    & -                   & -         \\
MABAS$^\star$ \cite{kim2020model} (ECCV,2020) & ResNet12
& {\color{blue}$\textbf{73.24} \pm 0.95 $}    & {\color{blue}$\textbf{85.65} \pm 0.65 $}    & {\color{blue}$\textbf{41.74} \pm 0.73 $}   & {\color{blue}$\textbf{57.11} \pm 0.75 $}   \\
\midrule
\textbf{Our MHFC}$^{\star\star}$ & ResNet12
& {\color{red}$\textbf{80.00} \pm 1.02 $}   & {\color{red}$\textbf{86.33} \pm 0.61 $}    & {\color{red}$\textbf{46.44} \pm 0.93 $}   & {\color{red}$\textbf{59.41} \pm 0.72 $}    \\
\midrule
\midrule
TEAM$^\dagger$ \cite{qiao2019transductive} (ICCV,2019) & ResNet12
& $70.43 \pm 1.03$    & $81.25 \pm 0.92$   & -                   & -    \\
Fine-tuning$^\dagger$ \cite{dhillon2020baseline} (ICLR,2020) & WRN
& {\color{blue}$\textbf{76.58} \pm 0.68 $}    & {\color{blue}$\textbf{85.79} \pm 0.50 $}    & {\color{blue}$\textbf{43.16} \pm 0.59 $}   & {\color{blue}$\textbf{57.57} \pm 0.55 $}   \\
\midrule
\textbf{Our MHFC}$^\dagger$ & ResNet12
& {\color{red}$\textbf{81.83} \pm 1.16 $}   & {\color{red}$\textbf{89.27} \pm 0.63 $}    & {\color{red}$\textbf{47.02} \pm 1.05 $}    & {\color{red}$\textbf{61.88} \pm 0.80 $}    \\
\midrule
\midrule
ICI$^\ddagger$ (80) \cite{wang2020instance} (CVPR,2020)  & ResNet12
& {\color{blue}$\textbf{78.07} $}            & {\color{blue}$\textbf{84.76}$}             & -                   & -    \\
\midrule
\textbf{Our MHFC}$^\ddagger$ ($80$) & ResNet12
& $84.74 \pm 1.14$   & $90.19 \pm 0.63$    & {\color{red}$\textbf{ 50.95} \pm 1.11 $}    & $ 64.05 \pm 0.82 $    \\
\textbf{Our MHFC}$^\ddagger$ ($100$) & ResNet12
& {\color{red}$\textbf{85.76} \pm 1.08 $}   & {\color{red}$\textbf{90.42} \pm 0.64$}    & $50.72 \pm 1.09$    & {\color{red}$\textbf{64.45} \pm 0.83$ }   \\
\bottomrule
\end{tabular}
}
\label{table: comparison_results_cifar_fc100} 
\end{center}
\end{table*}

\subsection{Implementation Details}
\label{subsec: Network Architectures}
In this paper, all the FEMs on different heads adopt the ResNet12 \cite{he2016deep} backbone, consisting of four residual blocks ($3\times 3$ convolution layer, batch normalization layer, LeakyReLU layer), four $2\times 2$ max pooling layers, and four dropout layers. 
We adopt stochastic gradient descent (SGD) optimizer with Nesterov momentum ($0.9$) for the optimizer.
For the parameter $\eta$ in \textbf{Equation \ref{eqa: obj_compute_Omega}}, we fix it to $1.4$ for convenience.
We set the training epochs to $120$ and test over $600$ episodes with $15$ query samples per class for all the models.
Besides, the selected subspace learning methods all follow the default implementation of scikit-learn \cite{pedregosa2011scikit}.
And there has no fine-tuning process when classifying the novel data.
For other settings, such as the learning rate, data augmentation, filters' number, we follow the ICI \cite{wang2020instance}.

\subsection{Experimental Results}
\label{subsec: Experimental Results}
We compare the proposed MHFC (only fuse SS-F-Fea and SS-M-Fea) with several state-of-the-art methods, the results are listed in \textbf{Table \ref{table: comparison_results_mini_tiered}} and \textbf{\ref{table: comparison_results_cifar_fc100}}. 
Here, we list several observations.

\textbf{(1)} 
Researchers split the few-shot learning methods into three settings, e.g., inductive, transductive, and semi-supervised. While, in the related works of FSL, the results on all settings are usually compared together.
Compared with the methods proposed recently, our MHFC has achieved state-of-the-art performance. It has far surpassed other models on the four datasets, especially on $5$-way $1$-shot case, the MHFC outperforms other methods at least \textbf{4.4\%}, \textbf{2.1\%}, \textbf{7.7\%} and \textbf{7.8\%} on mini-ImageNet, tiered-ImageNet, CIFAR-FS, FC100 datasets.
Our results on the $5$-way $1$-shot case are even better than many other methods on the $5$-way $5$-shot case.
And on the $5$-way $5$-shot case, the MHFC also exceeds others at least \textbf{3.0\%}, \textbf{2.8\%}, \textbf{4.8\%} and \textbf{6.9\%} on mini-ImageNet, tiered-ImageNet, CIFAR-FS, FC100 datasets.

\textbf{(2)}
Compared with the methods of each setting, our proposed MHFC has achieved excellent performances,
in particular on the $5$-way $1$-shot case with the inductive setting.
MHFC has significant improvements of at least \textbf{7.7\%}, \textbf{10.0\%}, \textbf{6.8\%} and \textbf{4.7\%} on mini-ImageNet, tiered-ImageNet, CIFAR-FS, FC100 datasets.
Notably, the performance of MHCF with the inductive setting has exceeded many other methods with transductive or semi-supervised settings.
Besides, for the methods based on semi-supervised setting, the final results are influenced by the number of employed unlabeled samples. 
Thus, we observe the impact and list the results in \textbf{Appendix B.1}. 
With the increase of unlabeled samples, the proposed method has become more effective. And the results start to saturate after $100$ unlabeled samples.


\begin{figure}
	\begin{center}
		\includegraphics[width=0.9\linewidth]{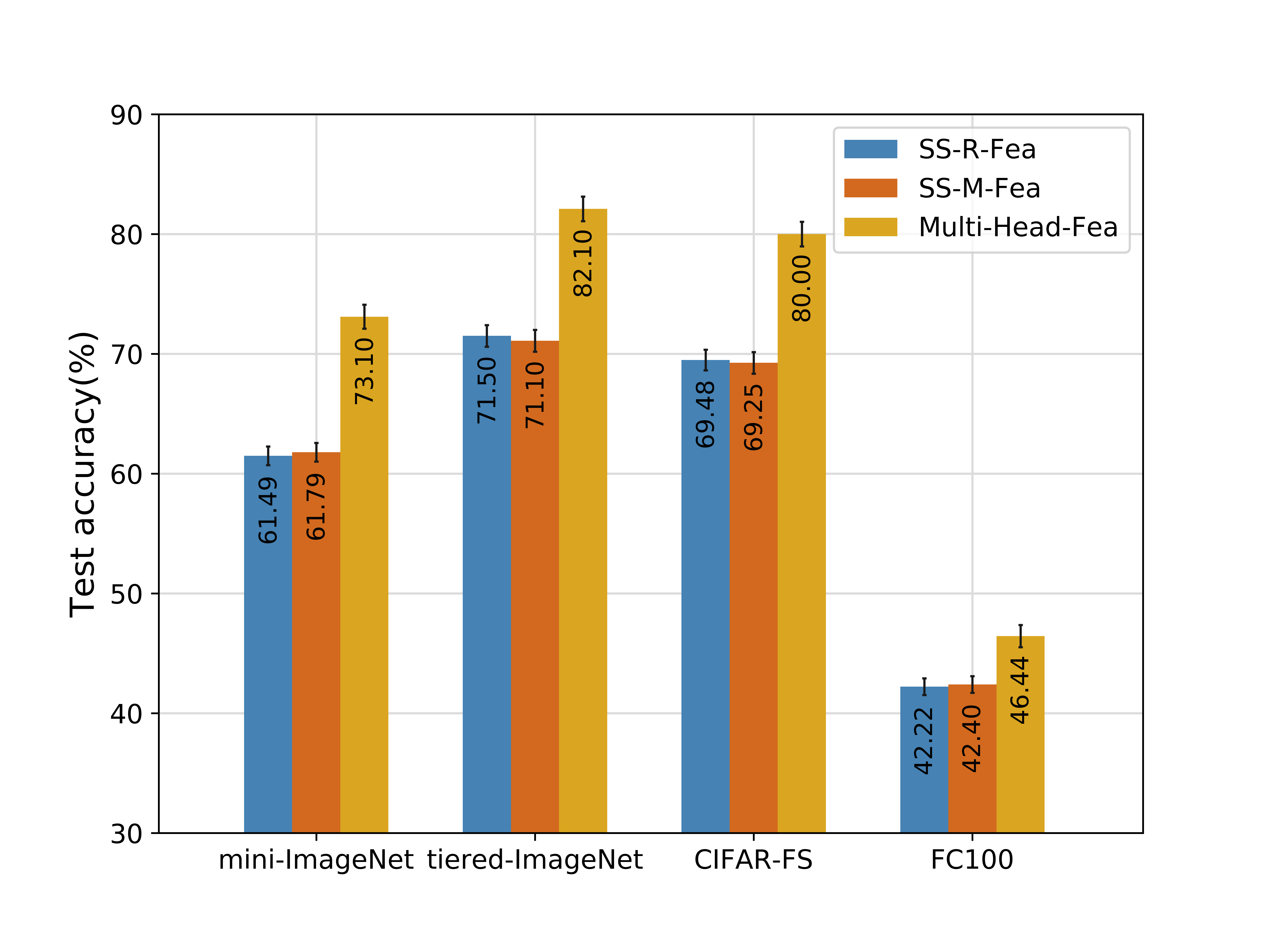}
	\end{center}
	\caption{Ablation studies to show the performances of employing different kinds of features with the inductive setting.}
	\label{fig: ablation_studie_multi_head_fea}
\end{figure}

\subsection{Ablation Studies}
\label{subsec: Ablation Studies}
In this paper, we propose MHFC for few-shot learning. 
There exist three factors influence the classification performance, e.g., \textbf{(1)} introducing multi-head features; \textbf{(2)} transforming the multi-head features into a unified space; \textbf{(3)} designing attention mechanism to weight features. 
This section conducts ablation studies on $5$-way $1$-shot case with the inductive setting to evaluate the efficiency of the three blocks.

\subsubsection{\textbf{Influence of Multi-Head Feature}}

The results listed in \textbf{Table \ref{table: comparison_results_mini_tiered}, \ref{table: comparison_results_cifar_fc100}} employ two heads of features (SS-R-Fea and SS-M-Fea). 
Here, we compare the results of employing the raw single-head feature with the collaborative feature (e.g., multi-head-fea).
The results of four datasets on $5$-way $1$-shot case with inductive setting are listed in \textbf{Figure \ref{fig: ablation_studie_multi_head_fea}}. Obviously, employing the collaborative feature improves significantly compared with exploiting the raw single-head feature. It demonstrates the efficiency of our "multi-head feature" to some extent.
Moreover, the proposed model can integrate more than two heads of features. We list the comparison results in \textbf{Appendix B.2}.


\begin{figure}
	\begin{center}
		\includegraphics[width=0.9\linewidth]{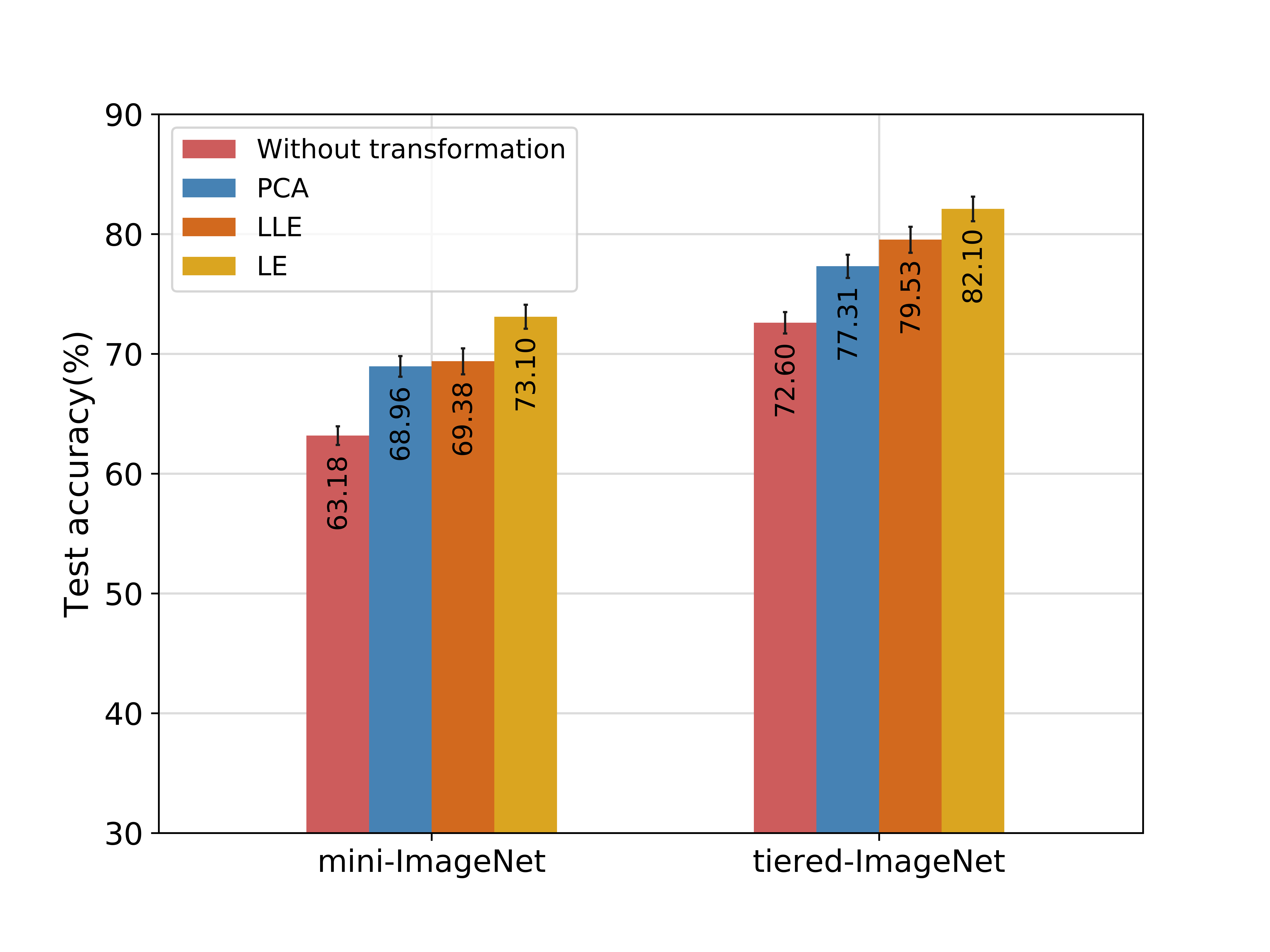}
	\end{center}
	\caption{The comparison results of adding transformation or not on $5$-way $1$-shot case with the non-standardized inductive setting.}
	\label{fig: ADD_LE}
\end{figure}

\subsubsection{\textbf{Influence of Subspace Transformation}}
For all the results listed in \textbf{Table \ref{table: comparison_results_mini_tiered}, \ref{table: comparison_results_cifar_fc100}} with inductive setting, we select LE \cite{belkin2002laplacian} to transform the multi-head features. 
While, besides the LE, there also exist several other choices to pull multi-head features to a unified space, such as PCA \cite{tipping1999probabilistic}, LLE \cite{roweis2000nonlinear}. We illustrate \textbf{Figure \ref{fig: ADD_LE}} to show the comparison results. We can see that all kinds of subspace learning methods are helpful to our MHFC, and LE is the most.

\begin{table}[!t]
\begin{center}
\caption{Comparison results with fixed weights on $5$-way few-shot case. (a, b) denotes that the SS-R-Fea's weight is "a", and SS-M-Fea's weight is "b". MHFC employs our designed attention block to update the weights automatically for each episode. The top two results are shown in red and blue.}
\setlength{\tabcolsep}{1.5mm}{
\begin{tabular}{lcccccc}
\toprule 
\multicolumn{1}{l}{\multirow{2}{*}{\textbf{Weight}}}
& \multicolumn{2}{c}{\textbf{mini-ImageNet}} 
& \multicolumn{2}{c}{\textbf{tiered-ImageNet}}
\\ 
\cmidrule(l){2-5}
\multicolumn{1}{c}{}
& \textbf{$1$-shot}   & \textbf{$5$-shot}  & \textbf{$1$-shot} & \textbf{$5$-shot}  \\  
\midrule
(0.1, 0.9)  
& $70.85$             & $80.09$            & $79.66$           & $87.12$   \\
(0.3, 0.7)  
& $71.32$          & {\color{blue}$\textbf{80.72}$}            & $80.25$           & $87.14$   \\
(0.5, 0.5)  
&{\color{blue}$\textbf{71.88}$}             & $80.15$          & {\color{blue}$\textbf{80.86}$}           & {\color{blue}$\textbf{87.38}$}   \\
(0.7, 0.3)  
&$70.56$           & $80.48$            & $80.19$           & $87.19$   \\
(0.9, 0.1)  
&$70.36$      & $79.97$            & $79.69$           & $86.93$   \\
\midrule
\textbf{MHFC$^{\star\star}$} 
& {\color{red}$\textbf{73.10}$}   & {\color{red}$\textbf{81.75}$}   & {\color{red}$\textbf{82.10}$}  & {\color{red}$\textbf{87.99}$}   \\
\bottomrule
\end{tabular}
}
\label{tab: fix_weight} 
\end{center}
\end{table}

\subsubsection{\textbf{Influence of Attention Block}}
We can use multi-head features to describe a category of samples, but the degrees of importance are different on each head.
To this end, it is crucial to design the attention block to calculate weights for different categories automatically.
Here, we compare the results with fixed weights to our MHFC, which is listed in \textbf{Table \ref{tab: fix_weight}}.
The results show that the updated weights are more reasonable for our method. 
Besides, from \textbf{Equation \ref{eqa: obj_compute_Omega}}, we know that $\eta$ is a parameter to influence the to-be-learned weights. For fairness and convenience, we have fixed $\eta$ to $1.4$ for all the experiments. We list the other comparison results in \textbf{Appendix B.3}.

\subsection{Cross-Domain Few-Shot Learning}
\label{subsec: Cross-Domain}
After introducing multi-head features from different views, we believe that the MHFC is an extremely robust method in practical scenarios. Therefore, we evaluate the proposed method with transductive setting on a cross-domain dataset: e.g., mini-ImageNet $\longrightarrow$ CUB. 
In pre-train stage, we use mini-ImageNet to train the FEM, and in meta-test stage, we classify the CUB dataset.
The results are reported in \textbf{Table \ref{table: comparison_results_cross-domain}}. 
Compared to the state-of-the-arts, we have significant improvements at least \textbf{13.3\%} on $1$-shot case and \textbf{7.8\%} on $5$-shot case.
Thus, the performance on the cross-domain few-shot learning task demonstrates that the MHFC can solve the DSP better, and the proposed MHFC would be powerful in real practice.

\begin{table}[!t]
\begin{center}
\caption{Comparison in cross-domain dataset scenario. Our MHFC is in transductive setting. $(\cdot)^\flat$ and $(\cdot)^\sharp$ indicate the reported results come from \cite{boudiaf2020transductive} and  \cite{mangla2020charting}, respectively. The top two results are shown in red and blue.}
\setlength{\tabcolsep}{2.0mm}{
\begin{tabular}{lcccccc}
\toprule 
\multicolumn{1}{l}{\multirow{2}{*}{\textbf{Method}}}
& \multicolumn{2}{c}{\textbf{mini-ImageNet $\longrightarrow$ CUB}} 
\\ 
\cmidrule(l){2-3}
\multicolumn{1}{c}{}
& \textbf{$5$-way $1$-shot} & \textbf{5-way 5-shot}  \\  
\midrule
Baseline$^\flat$ \cite{chen2019closer}
& -                  & $53.1$    \\
ProtoNet$^\flat$ \cite{snell2017prototypical}
& -                  & $62.0$    \\
RelationNet$^\flat$ \cite{sung2018learning}
& -                  & $57.7$    \\
GNN$^\flat$ \cite{tseng2020cross}
& -                  & $66.9$    \\
Neg-Cosine$^\flat$ \cite{liu2020negative}
& -                  & $67.0$    \\
LaplacianShot$^\flat$ \cite{ziko2020laplacian}
& -                  & $66.3$    \\
TIM-GD$^\flat$ \cite{boudiaf2020transductive}
& -                  & {\color{blue}$\textbf{71.0}$}    \\
\midrule
MetaOpt$^\sharp$ \cite{bertinetto2019metalearning}
& $44.79 \pm 0.75$   & $64.98 \pm 0.68$    \\
Manifold Mixup$^\sharp$ \cite{verma2019manifold}
& $46.21 \pm 0.77$   & $66.03 \pm 0.71$    \\
S2M2$^\sharp$ \cite{mangla2020charting}
& {\color{blue}$\textbf{48.24} \pm 0.84$}   & $70.44 \pm 0.75$    \\
\midrule
\midrule
\textbf{MHFC}$^\dagger$
& {\color{red}$\textbf{61.57} \pm 1.28$} & {\color{red}$\textbf{78.80} \pm 0.78$}\\
\bottomrule
\end{tabular}
}
\label{table: comparison_results_cross-domain} 
\end{center}
\end{table}

\section{Conclusion}
Few-shot learning (FSL) based tasks have a fundamental problem, e.g., distribution-shift-problem (DSP). 
To address this challenge, we propose Multi-Head Feature Collaboration (MHFC), which attempts to collaboratively represent samples by fusing multi-head features.
It is helpful to strengthen the FSL based model’s efficacy and robustness.
MHFC is a simple non-parametric method that can directly employ the existing FEMs.
Experimental results have demonstrated the effectiveness of MHFC.

\section*{Acknowledgment}
The paper was supported by 
the National Natural Science Foundation of China (Grant No. 62072468), 
the Natural Science Foundation of Shandong Province, China (Grant No. ZR2019MF073), 
the Fundamental Research Funds for the Central Universities, China University of Petroleum (East China) (Grant No. 20CX05001A),
the Graduate Innovation Project of China University of Petroleum (East China) (YCX2021117, YCX2021123).

\bibliographystyle{ACM-Reference-Format}
\balance
\bibliography{Ref}

\end{document}